\documentclass{elsart}               

\usepackage{graphicx}          
 
 \usepackage{amsmath}      
 \usepackage{amssymb}
\usepackage{amsbsy}  
 \usepackage[nolist]{acronym}
\usepackage{savesym}
\savesymbol{AND}
\usepackage{algorithmic}
 \usepackage{xcolor}
\begin{document}
\pdfminorversion=4


\begin{frontmatter}

\title{Dual Control for Exploitation and Exploration (DCEE) in Autonomous Search\thanksref{EPSRC}}
\thanks[EPSRC]{This paper has not been presented on any IFAC conference. Corresponding author: Wen-Hua Chen. w.chen@lboro.ac.uk}

\date{\empty}

\author[WHC]{Wen-Hua Chen} \ead{W.Chen@lboro.ac.uk}
\author[WHC]{Callum Rhodes} \ead{C.Rhodes@lboro.ac.uk}
\author[WHC]{Cunjia Liu}  \ead{C.Liu5@lboro.ac.uk}
\address[WHC]{Department of Aeronautical and Automotive
  Engineering\\
Loughborough University\\
Leicestershire, LE11 3TU\\
UK }


\date{\empty}
\maketitle

\begin{abstract}
This paper proposes an optimal autonomous search framework, namely Dual Control for Exploration and Exploitation (DCEE), for a target at unknown location in an unknown environment. Source localisation is to find sources of atmospheric hazardous material release in a partially unknown environment. 
This paper proposes a control theoretic approach to this autonomous search problem. To cope with an unknown target location, at each step, the target location is estimated by Bayesian inference. Then a control action is taken to minimise the error between future robot position and the hypothesised future estimation of the target location. The latter is generated by hypothesised measurements at the corresponding future robot positions (due to the control action) with the current estimation of the target location as a prior. It shows that this approach can take into account both the error between the next robot position and the estimate of the target location, and the uncertainty of the estimate. This approach is further extended to the case with not only an unknown source location, but also an unknown local environment (e.g. wind speed and direction). Different from current information theoretic approaches, this new control theoretic approach achieves the optimal trade-off between exploitation and exploration in a unknown environment with an unknown target by driving the robot moving towards estimated target location while reducing its estimation uncertainty. This scheme is implemented using particle filtering on a mobile robot. Simulation and experimental studies demonstrate promising performance of the proposed approach. The relationships between the proposed approach, informative path planning, dual control, and classic model predictive control are discussed and compared. This work opens a door for further developing control systems operating in unknown environments, or performing tasks with unknown parameters.
\end{abstract}

\end{frontmatter}


\begin{acronym}
    \acro{MPC}{model predictive control}
    \acro{STE}{source term estimation}
    \acro{IP}{Isotropic plume}
    \acro{UAV}{unmanned aerial vehicle}
    \acro{RMSE}{root mean square error}
    \acro{PDF}{probability density function}
    \acro{PID}{photoionisation detector}
    \acro{DWA}{dynamic window approach}
    \acro{SLAM}{simultaneous localisation and mapping}
\end{acronym}


\section{Introduction}
\label{sec:intro}
Searching sources of airborne substance release could find a wide range of applications from disaster management and environment protection, to gas leakage detection in oil and gas industry \cite{Singh2015}, \cite{hutchinson2017review}. Chemical and biological materials could be released into the atmosphere deliberately (e.g. discharged by a plant, terrorism), naturally  (e.g. methane emission, volcanic eruption), or accidentally (e.g. accidents in a chemical plant, Fukushima nuclear disaster). It can also be seen in natural behaviours from animals searching for food sources based on odours, to insects seeking mating in a large field or forest \cite{vergassola2007infotaxis}. Quite often source localisation and quantification is referred to as a source term estimation (STE) problem in literature \cite{bieringer2015automated}, \cite{platt2010comparative}, \cite{hutchinson2017review}. In the presence of hazardous material release events, there are two main approaches for existing emergency response practices: a static network of pre-deployed sensors and the manual collection of sensor measurements, e.g. using hand held devices and dedicated manned vehicles. The former is costly and requires substantial pre-planning while the latter puts people in harms way. After data are collected, estimation algorithms can be utilised to estimate the location of sources and release rates.

With the advances of robotics and autonomous system technologies, there is a strong interest in developing a new STE solution using recently available mobile sensor platforms, where chemical or biological sensors are installed on a mobile ground robot or an unmanned aerial vehicle (UAV) \cite{hutchinson2017review}. Essentially, this is to search a source in an unknown environment. It could be conducted through exhaustive search, so it can be considered as a coverage search problem. However, due to the time critical nature of this type of mission, there is a strong need in speeding up the search process, particularly for searching in a large area. To this end, various tools have been introduced including optimisation and Bayesian approaches \cite{hutchinson2017review}. A key challenge arising in autonomous search using a mobile sensor platform is the \emph{deep interaction between estimation and planning} -- based on the current belief, a decision shall be made as to where to take the next measurement in order to maximise the chance of finding the source which, in turn, changes the belief since the new measurement is obtained at a new location. Driven by information-theoretic approaches, informative path planning (IPP) offers a promising solution to the STE problem \cite{bostrom2019informative}. In an IPP approach, a mobile sensor platform plans a path that maximises the information gain about the source and local environment, based on its current belief, and updates its belief through incorporating new measurements of the dispersion environment using on-board sensors  \cite{ristic2010information} \cite{hutchinson2019use}. A number of cognitive search strategies have been developed based on different reward functions of the information gain; for example, \cite{ristic2017autonomous} \cite{hutchinson2018entrotaxis} \cite{HutchinsonICRA2018}.  A complete autonomous search system has been developed and tested in both indoor and outdoor environments with ground robots and unmanned aerial vehicles, respectively \cite{hutchinson2019unmanned} \cite{hutchinson2018information}.  In all the above studies, source localisation and search is considered as an estimation problem, or an exploration problem of unknown environment, so information theoretic approaches were advocated.  

In this paper, instead of taking the information theoretic approach, we reconsider this autonomous search problem from a control theoretic perspective, and propose a completely new framework. Finding an unknown source is interpreted as \emph{controlling} a mobile sensor platform approaching a target.  That is, our goal is to design a control system that autonomously drives a robot equipped with chemical sensors to approach a target at an \emph{unknown location} in an \emph{unknown environment}. Different from a classic control setting, there is no reference trajectory or prescribed setpoint for the robot to follow. Indeed, the target location is unknown and the path to the target is to be defined in this autonomous search problem. To solve this problem, we formulate it as a stochastic model predictive control (MPC) problem with an undefined target location. It is shown that our approach is not only intuitive and promising, but also closely links with the dual control concept \cite{Fel1960a} \cite{Fel1960b}, \cite{heirung2015mpc}, \cite{Mes2018}.

Different from a classic control setting, dual control considers that a control action affects not only future states of a system, but also uncertainty of its estimation \cite{BarTse1974}. 
In general, dual control is intractable and computationally expensive, therefore it has found very few practical applications (\cite{Wit2008}). It shall be noted that there is a significant difference between the existing dual control formulation and our approach. In the current dual control setting, dual effects consider the effect of control on systems (e.g. state estimation in stochastic control, e.g.\cite{Mes2018}, or parameter estimation in adaptive control, e.g. \cite{Wit2008}, \cite{FilUnb2000}, \cite{heirung2015mpc}). The dual control effect we consider in this paper is about the probing effect on the operational environment or the target, rather than on the uncertainty of the dynamic system itself (e.g. state or parameters). We aim to drive the robot to a believed target position (to be defined later) but at the same time reduce the uncertainty associated with this believed target position, and build up a better understanding of the operational environment. This very difference gives us the ability to deal with the problem of controlling a robot approaching a target at an \emph{unknown} location and in an \emph{unknown} environment.


The problem we are trying to solve also belongs to a much broader class of problems arising in machine learning and artificial intelligence - the trade-off between \emph{exploitation} and \emph{exploration} in an unknown environment. In an unknown environment, \emph{exploitation} makes best choice based on current information which is biased to what you have known and may lead to local results. In contrast, \emph{exploration} aims to gather more information and make the best overall choice, but may lead to wasted efforts. In our view, IPP in robotics and autonomous systems focuses on exploration to increase information gain so to reduce uncertainty, while traditionally control engineering mainly focuses exploitation, i.e. making use of information to derive a suitable control action/plan. It is shown that for an autonomous search of the source of airborne substance release, our approach provides an optimal trade-off between exploitation and exploration in the sense of Bellman's principle of optimality (1957) \cite{Bellman1957Dynamic}.



This paper first formulates the autonomous search problem in an unknown environment into a control problem. The cost function is defined as the expected error between the robot's future position and the \emph{predicted} estimation of the target location (after taking into account information gain due to hypothesised future measurements). Since the true chemical concentration measurements at future positions of the robot are unknown, this predicted estimation is generated by a dispersion model with the current estimated source location as a prior and randomly simulating sensor readings under described sensor characteristics. It is shown that the cost function actually consists of two parts: the first part is about driving the robot moving towards the \emph{predicted} estimated source location with hypothesised measurements, while the second term quantifies the uncertainty level of the \emph{predicted} estimated source location. In other words, the first term is related to \emph{exploitation} by making good use of the estimated source location in generating control actions, whereas the second term is about \emph{exploration} by using control to probe the unknown environment in order to reduce the uncertainty level of the location estimation and the belief of the environment. By minimising this cost function, we are able to optimally trade-off between these two effects, i.e. simultaneously driving the robot towards the source and reduce the uncertainty in source location estimation. It is pointed out that, to a large extent, the current MPC and information theoretic approach could be considered as special cases of our framework. 

Bayesian inference is used to develop the control algorithm. The whole control framework consists of two steps: Bayesian estimation of the source location (and other associated environmental parameters) after a new chemical concentration measurement is taken, and the calculation of a control action by minimising the above cost function where Bayesian estimation is also embedded for calculating the predicted posteriors. However, it is difficult to implement this Bayesian estimation based control framework, particularly when a large number of source and environment related parameters are unknown. We resort to particle filtering for implementing the proposed control algorithm in both simulation and experimental tests.

 There are 4 main contributions in this paper. 1). This paper formulates the autonomous search of airborne hazardous substance release as a new type of control problem from a control theoretic perspective. The main feature of this control problem is that the target is at an unknown location and there is also no predefined reference trajectory.  In other words, the goal of the control system is well defined, i.e. finding the source, but the information about the target and its operational environment is \emph{incomplete}. 2). Inspired by a dual control concept, a new framework is proposed to trade-off between making use of the current estimate to driving a robot towards the believed location of the target (exploitation) and reducing the uncertainty of the estimation (exploration), namely Dual Control for Exploration and Exploitation (DCEE). It is shown that optimal autonomous search is realised through this approach (in the sense of Bellman optimality \cite{Bellman1957Dynamic}). This work is extended to deal with an unknown environment. 3). Simulation and physical experiments have been developed to demonstrate the performance of the proposed DCEE. Numerical implementation of Bayesian estimation and online optimisation have been presented. The comparison with alternative approaches including information theoretic approaches and classic MPC clearly shows superior performance and effectiveness of the proposed DCEE. 4). This work brings insights into a number of related areas such as information theoretic approaches (e.g. IPP, active sensing), dual control, exploration and exploitation, autonomous search, and MPC. It shows that the existing approaches are biased to  (or only focus on) either exploitation or exploration so may lead to less efficient or poorer searching performance.      

The remainder of the paper is organised as follows. Section~\ref{sec:formulation} introduces the autonomous search problem for atmospheric dispersion of hazardous substances. A isotropic plume dispersion model is presented and chemical sensor behaviour with misdetection and sensor errors are also introduced. The control algorithm development is presented in Section~\ref{sec:CAD}. An one-step-ahead cost function is proposed and the control algorithm is derived with discussions. Particle filtering implementation of the proposed control algorithm is presented to facilitate real-time applications. This work is further extended in several directions in Section~\ref{sec:extension} including operating in an unknown environment (for example, wind speed and direction has a significant effect on the dispersion) and a multi-stage cost function. The relationship between DCEE and other relevant control/search strategies is also discussed in this Section to provide more insight. Simulation is conducted in Section~\ref{sec:simulation} and comparison with IPP and MPC clearly demonstrates superior performance of the proposed framework. Furthermore, physical experiments have been conducted on a ground robot in an indoor environment in Section~\ref{sec:exp}. This paper ends with conclusion in Section~\ref{sec:conclusion}.

\section{Autonomous search and its formulation} \label{sec:formulation}

\subsection{Agent modelling}
Consider an autonomous agent (e.g. a robot or a UAV) is tasked to search an area for finding a possible source of airborne hazardous substance release. It is supposed that there is a lower level controller that drives the agent to following any planned path so the detailed dynamics of the agent are ignored for the path planning purpose. The behaviour of the agent is modelled as \cite{hutchinson2019use}
\begin{equation} \label{eq:agent}
   \mathbf{p}_{k+1}=\mathbf{p}_{k}+\mathbf{u}_{k}, 
\end{equation} 
where $\mathbf{p}_k \in R^3 $ is the current position of the agent, $\mathbf{u}_{k} \in \mathcal{U}$ is the movement of the agent and $\mathcal{U}$ is a set of admissible control actions.

\subsection{Dispersion model and environment}
The dispersion model of an airborne chemical release can be formulated as an Isotropic plume model \cite{holmes2006review}. In this model, the expected concentration at the robot position $\mathbf{p}_k = \left[ p_{x,k}, p_{y,k}, p_{z,k} \right]^\mathsf{T}$ from a release source positioned at $\mathbf{s} = \left[ s_{x}, s_{y},s_{z} \right]^\mathsf{T}$ with a releasing rate of $q_s$ is given by:
	\begin{multline}\label{eq:advdif}
		\mathcal{M}\left(\mathbf{p}_k,\Theta \right) = \frac{q_s}{4\pi \zeta_{s1}||\mathbf{p}_k-\mathbf{s}||}  \exp \left[\frac{-||\mathbf{p}_k-\mathbf{s}||}{\lambda}\right] \times \\ \exp \left[\frac{-(p_{x,k}-s_{x})u_s \cos\phi_{s}}{2 \zeta_{s1}}\right]\exp \left[\frac{-(p_{y,k}-s_{y})u_s \sin\phi_{s}}{2 \zeta_{s1}}\right],
	\end{multline}
where $\lambda = \sqrt{\frac{\zeta_{s1} \zeta_{s2}}{1+ (u_s^2 \zeta_{s2})/(4 \zeta_{s1})}}$. A number of parameters are involved to characterise this dispersion, including the average particle lifetime $\zeta_{s2}$, the diffusivity $\zeta_{s1}$, the mean wind speed $u_s$ and wind direction $\phi_{s}$. These parameters of the source and local environment (e.g. wind field, temperature) are assumed be unknown but may have certain prior information. 

\subsection{Sensor modelling}
 When an agent moves to a location, a point-wise measurement of chemical/biological concentration at the current location is taken by onboard chemical (or biological) sensors/detectors. Due to the nature of these sensors, an agent needs to hold at the location for a short period in order to get a reliable reading, which is referred to as the sampling interval. This is also why the detailed dynamics of an agent are not required in plan planning. Due to the power, size and payload of mobile platforms (e.g. small UAVs), only portable chemical detectors could be installed on the agent which quite often implies a limited accuracy and poor dynamic measurement capability. Furthermore, local turbulence in airflow has a significant effect on the distribution of concentration, causing quick perturbation in the dispersion, which causes intermittent sensor readings. Therefore, it is important to model sensor characteristics and the influence of the local environment on its behaviour. Both detection and non-detection events are considered due to the sparsity of the measurements and complication in local flow. A non-detection event may be caused by three hypothesised scenarios:
\begin{itemize}
    \item The source is not present at all or present but not within range of the chemical detector. Any concentration measurement recorded is only a result of background and instrument noise. 
    \item  A source is present and in detectable concentrations but there is a non-detection as
a result of intermittency caused by turbulence or a missed detection, typically exacerbated by the short sampling intervals of the agent.
\item The source is present but the concentration (plus any addition from background and
instrument noise) is below a pre-specified concentration threshold $z_{thr}$. The threshold is set high enough to minimise false detections, whilst maintaining sufficient sensitivity \cite{yee2017automated}. 
\end{itemize}      

In a detection event, the sensor reading consists of the true concentration and sensor noises.    
In summary, the sensor behaviour is modelled as
\begin{equation} \label{eq:sensor}
    z_k=\left \{ \begin{array}{ll}\mathcal{M}(\mathbf{p}_k,\Theta)+v_k; &  D=1 \\ \bar{v}_k; & D=0 \end{array} \right.
\end{equation}
where $D$ denote a detection event. $P_d = Pr\{D = 1\}$ is the detection probability encapsulating all the three hypothesised scenarios. A Poisson distribution is used  to represent the non-detection event in this study with its non-detection probability as the sum of all these three scenarios. $\bar{v}_k$ and $v_k$ satisfying Gaussian distribution represent the noise in a non-detection event and detection event, respectively.   

\subsection{Autonomous search as a control problem}

Traditionally the source search problem is referred to as an STE problem. This is to estimate the key parameter related to release so emergency responders or disaster management teams could be well informed. Broadly speaking, there are two approaches in developing information theoretic based search strategies to this problem: optimisation based approaches where gradients of the concentration are exploited and Bayesian approaches which is based on a probability framework \cite{hutchinson2017review}. Various information metrics such as mutual information on entropy, Kullback-Leibler divergence and variance have been used as a reward function to develop cognitive search strategies such as `infotaxis'  \cite{ristic2017autonomous}, \cite{ristic2016study}, `entrotaxis' \cite{hutchinson2018entrotaxis}. They provide a more reliable and robust solution with a shorter search time than the optimisation based search methods and traditional search methods such as `chemotaxis' \cite{adler1966chemotaxis}. In essence, these information theoretic methods mainly focus on exploration, i.e., to explore the unknown environment to find more information about sources, although some inherent trade-off behaviour between exploitation and exploration has been observed since it also drives the agent towards the target \cite{hutchinson2018information}.  

Departing from the information theoretic approaches, we reformulate this problem from a control theoretic perspective. Basically, we consider the autonomous search problem as a process to driving the agent from its start location to a target. Once the target is reached, the mission is completed. However one challenge is that this is a non-conventional control problem since the location of the target is unknown and the environment is also unknown. It cannot be formulated in the traditional control framework such as as a regulation or tracking problem as there is no pre-defined reference trajectory. Another challenge from the control system point of view is that measurements are sporadic and intermittent. In actuality, for the majority of the search process, there is no sensor reading. To best of the authors' knowledge, there is no work in investigating this kind of autonomous search from a control theoretic perspective. We will show our new formulation provides a much better way to balance exploitation (make use of information collected) and exploration (reduce uncertainty in information by exploring the environment) as to lead to a substantially improved search performance.


\section{Control algorithm development} \label{sec:CAD}

We will start control algorithm development with only unknown parameters related to the source but will extend the work to deal with an unknown environment in Section~\ref{sec:extension}. Consequently, the unknown or uncertain parameters we consider at this stage are the location of the source and its release rate, i.e.  $\Theta = \begin{bmatrix}\mathbf{s}^{\mathsf{T}} & q_s \end{bmatrix}^\mathsf{T}$.

The goal of the control system for autonomous search is to drive the agent to the source at an unknown location. Up until now, a typical approach in the current control system setting would be to drive the robot towards the believed location derived from prior information and collect measurements. However, this would not be able to take into account the current level of uncertainty in the belief and how the future move could reduce it as illustrated in Section~\ref{sec:MPC}. This motivates our DCEE framework.

\subsection{DCEE framework}

When the source term $\Theta$ is unknown, a probability density function $p(\Theta_k)$ can be used to represent the belief of $\Theta$ at time $k$. Let $\mathcal{Z}_{k}$ denote the vector of measurements collected at different locations by the robot up to time step $k$, i.e.
$\mathcal{Z}_{k}:=\{z_{1}(\mathbf{p}_{1}),z_{2}(\mathbf{p}_{2}),\ldots, z_{k}(\mathbf{p}_{k})\}$. 

We define $\rho_{k|k}:=p(\Theta|\mathcal{Z}_{k})$ as the posterior distribution at time $k$. Now we consider any control input $\mathbf{u}_k$, which moves the robot to a new location where a future new measurement $\hat{z}_{k+1}$ could be made at time $k+1$ (where $\hat{z}$ is used to distinguish a predicted variable from a real variable). This future measurement can be considered as a random variable given the control input, i.e. $\hat{z}_{k+1} \sim p(\hat{z}_{k+1}|\mathbf{u}_{k})$. We define a hypothetical posterior distribution $\hat{\rho}_{k+1|k}$ conditioned on this possible future measurement $\hat{z}_{k+1}$, i.e. $\hat{\rho}_{k+1|k}:=p(\Theta|\mathcal{Z}_{k},\hat{z}_{k+1})$. Therefore, the control input not only affects the future location of the robot but also affects its future belief about where the target might be.  


Inspired by the above discussion, we would like to move the robot's future position closer to the predicted posterior of the target's location, $\hat{\rho}_{k+1|k}$, conditional upon the control input $\mathbf{u}_k$.  Given a possible future measurement $\hat{z}_{k+1}$ induced by control input $\mathbf{u}_{k}$, the conditional cost function for approaching a target at an unknown location $\mathbf{s}$, can be formulated as follows
 \begin{equation} \label{eq:DCEE}
J(\mathbf{u}_{k}) =\mathbb{E}_{\Theta} \left[ \mathbb{E}_{\hat{z}_{k+1}} \left[ \| \mathbf{p}_{k+1|k} - \mathbf{s} \|^2 | \mathcal{Z}_{k}, \hat{z}_{k+1} \right] \right]
 \end{equation}
In this formulation, the control input $\mathbf{u}_{k}$ not only determines the future robot position $\mathbf{p}_{k+1|k}$, but also affects the possible future measurement $\hat{z}_{k+1}(\mathbf{p}_{k+1|k})$. Let $\Theta_{k+1|k}=[\mathbf{s}_{k+1|k}^\mathsf{T}, q_{s,k+1|k}]^\mathsf{T}$ denote the estimate of the source location and the release rate at time $k$ with the measurements $\mathcal{Z}_{k}$ and the virtual measurement $\hat{z}_{k+1}(\mathbf{p}_{k+1|k})$. It is clear that the move $\mathbf{u}_k$ affects the predicted posterior $\hat{\rho}_{k+1|k}:=p(\Theta_{k+1|k})$ of the source location and the release rate. Taking the expectation with respect to the future measurement rewards the exploration effect in reducing the level of uncertainty in unknown source parameter estimation, as we will show later.  

The optimisation problem for DCEE can be formulated as follows: 
\begin{subequations}
 \label{eq:dualCost}
 \begin{align}
    \min_{\mathbf{u}_{k}} J(\mathbf{u}_{k}) &=  \min_{\mathbf{u}_k} \mathbb{E}_{\Theta} \left[ \mathbb{E}_{\hat{z}_{k+1}} \left[\| \mathbf{p}_{k+1|k} - \mathbf{s} \|^2 | \mathcal{Z}_{k+1|k} \right] \right] \\
    \text{subject to} \nonumber \\
    & \mathbf{p}_{k+1|k}=\mathbf{p}_{k}+\mathbf{u}_{k} \\
    & \mathbf{u_k} \in \mathcal{U}
\end{align}
\end{subequations}
where $\mathcal{Z}_{k+1|k}:=\{ \mathcal{Z}_{k}, \hat{z}_{k+1} \}$. 


We define the nominal estimated source location as the mean of the posterior distribution of the source location estimation. Similarly, the nominal predicted estimation of the source location $\bar{\mathbf{s}}_{k+1|k}$ is defined as the mean of the predicted distribution, i.e. $\hat{\rho}_{k+1|k}$,  with $\mathcal{Z}_{k+1|k}$, which is given by
 \begin{equation}
     \bar{\mathbf{s}}_{k+1|k}:=\mathbb{E} \left[ \mathbf{s}_{k+1|k} \right] = \mathbb{E} \left[\mathbf{s}| \mathcal{Z}_{k+1|k}\right ] 
 \end{equation}
 With $\bar{\mathbf{s}}_{k+1}$, we define $\tilde{\mathbf{s}}_{k+1|k}$ conditional on $\mathcal{Z}_{k+1|k}$ as $\tilde{\mathbf{s}}_{k+1|k} =\mathbf{s}-\bar{\mathbf{s}}_{k+1|k}$. Therefore, the cost function (\ref{eq:dualCost}) can be reformulated as 
 \begin{equation}
 \label{eq:dualCost2}
J(\mathbf{u}_{k}) = \mathbb{E}_{\Theta,\hat{z}_{k+1}} \left[\| \mathbf{p}_{k+1|k} - \bar{\mathbf{s}}_{k+1|k} - \tilde{\mathbf{s}}_{k+1|k}  \|^2 | \mathcal{Z}_{k+1|k} \right]   
 \end{equation}

\begin{thm} \label{thm1}
The cost function of DCEE defined in (\ref{eq:DCEE}) is equivalent to the following cost function
\begin{equation}
\label{eq:theorem1}
   J(\mathbf{u}_{k})= \| \mathbf{p}_{k+1|k} - \bar{\mathbf{s}}_{k+1|k}\|^2 + P_{k+1|k}
\end{equation}
where $P_{k+1|k}=\mathbb{E}_{\Theta,\hat{z}_{k+1}} \left[ \tilde{\mathbf{s}}_{k+1|k}^T\tilde{\mathbf{s}}_{k+1|k}  | \mathcal{Z}_{k+1|k} \right]$ is the predicted covariance matrix of $\mathbf{s}_{k+1|k}$.
\end{thm}
 
 \begin{pf}
Expanding the square and collecting terms of the right hand side of (\ref{eq:dualCost2}) gives
\begin{equation}
\begin{split}
    J(\mathbf{u}_{k})
    =& \mathbb{E}\left[ \| \mathbf{p}_{k+1|k} - \bar{\mathbf{s}}_{k+1|k} \|^2 | \mathcal{Z}_{k+1|k}  \right] \\
    & - 2 \mathbb{E}\left[ \tilde{\mathbf{s}}_{k+1|k}^{T} (\mathbf{p}_{k+1|k} - \bar{\mathbf{s}}_{k+1|k}) | \mathcal{Z}_{k+1|k}   \right] \\
    & + \mathbb{E} \left[ \tilde{\mathbf{s}}_{k+1|k}^T\tilde{\mathbf{s}}_{k+1|k}  | \mathcal{Z}_{k+1|k} \right] 
\end{split}
\end{equation}
 Given that both $\mathbf{p}_{k+1|k}$ and $\bar{\mathbf{s}}_{k+1|k}$ are deterministic and that $\mathbb{E}\left[ \tilde{\mathbf{s}}_{k+1|k}^{T} | \mathcal{Z}_{k+1|k}\right] = 0$, results in (\ref{eq:theorem1}) can be concluded. 
 \end{pf}
 
 
 \begin{rem}
 The cost function in the form of (\ref{eq:theorem1}) clearly reveals the dual control effect of our approach. The first term in the cost function (\ref{eq:theorem1}) drives the robot to the estimated location of the source, which is related to exploitation. The second term is about the level of uncertainty of the estimated target location, captured by the predicted covariance of the source location in the next time step. The influence of future control action on both the distance to a believed target location and the current information uncertainty is quantified. It is quite intuitive and natural. Optimising this cost function over admissible control actions gives the best next move that balances the probing effect and performing certain task.
 \end{rem}
 
 \begin{rem}
 This optimal control problem gives the best strategy in trade-off between exploration and exploitation so it is the optimal autonomous search strategy for this problem in the sense of the principle of optimality (Bellmman, 1957). The existing IPP based autonomous search is only concerned about the second term while the classic MPC approach only considers the first term. In many areas, there is a big challenge in how to balance exploration and exploitation, particularly in artificial intelligence, optimisation and decision making for complicated problems. Normally weights on the cost/reward functions have to be introduced to balance these two effects. Sometimes, in order to reflect the dual effect, a related term is also artificially added to the cost function \cite{heirung2015mpc}. There is no such a requirement in choosing weights to trade-off in our formulation. They are derived from a physically meaningful cost function in (\ref{eq:DCEE}). The balance between them is naturally embedded. More discussion about this approach and the existing work will be made in Section~\ref{sec:extension}.
  \end{rem}
 
In the following sections, we will discuss the implementation issues related to this framework.

\subsection{Implementation of Bayesian estimation} 

Bayesian estimation plays a key role both in the inference engine for estimating the parameters related to the source and in the planning loop for calculating predicted posteriors.      

The conditional probability of the source terms can be obtained via recursive Bayesian estimation, such that  
\begin{equation} \label{bayes}
p(\Theta_{k}|\mathcal{Z}_{k}) 
= \frac{p(z_{k}|\Theta_{k})p(\Theta_{k} |\mathcal{Z}_{k-1})} {p(z_{k}|\mathcal{Z}_{k-1})} 
\end{equation}
where
\begin{equation} \label{bayes2}
p(z_{k}|\mathcal{Z}_{k-1}) = \int p(z_{k}|\Theta_{k} )p(\Theta_{k} | \mathcal{Z}_{k-1}) \, \mathrm{d} \Theta_{k}.
\end{equation}

The initial prior distribution $p(\Theta_{0}|\mathcal{Z}_{0})=p(\Theta_{0})$ is assumed to be given. In this work, the parameters associated with the source term are assumed to be unknown but fixed. The likelihood function in the Bayesian estimation $p(z_{k}|\Theta_{k})$ is determined by using the dispersion model (\ref{eq:advdif}) and the sensor model (\ref{eq:sensor}).

Given the nonlinear nature of this Bayesian estimation problem, a particle filtering approach is used to approximate the Bayesian estimation. In the particle filter, the posterior distribution from Eq.\,\eqref{bayes} is approximated by a set of weighted random samples $\{\Theta_k^{(i)},w_k^{(i)}\}_{i=1}^{N}$ such that 
\begin{equation} \label{eq:pf}
p(\Theta_{k}|\mathbf{z}_{1:k}) \approx \sum_{i=1}^{N} w_k^{(i)} \delta(\Theta_{k} - \Theta_k^{(i)}),
\end{equation}
where $\delta(\cdot)$ is a Dirac delta function, $\Theta_k^{(i)}$ is a sample representing a potential source term realisation and $w_k^{(i)}$ is the corresponding normalised weighting such that $\sum_{i=1}^{N}  w_k^{(i)} = 1$.
	
The process of recursively calculating the posterior distribution at sampling instance $k$ is summarised in Algorithm \ref{alg:PF_STE} and more details can be found in \cite{hutchinson2018information}.
 
\begin{algorithm}[h]
	\begin{algorithmic}[1]
	\REQUIRE prior samples: $\{\Theta_{k-1}^{(i)}, \omega_{k-1}^{(i)}\}_{i=1}^{N}$; sensor measurement $z_{k}$ at location $\mathbf{p}_{k}$;
		\FOR{$i=1, 2, \ldots, N$}
		\STATE draw sample $\Theta_{k}^{(i)} \sim q(\Theta_{k-1}^{(i)})$
		\STATE Assign weight $\bar{w}_{k}^{(i)} = w_{k-1}^{(i)} \cdot \frac{ p (z_{k}|\Theta_{k}^{(i)}) p(\Theta_{k}^{(i)}|\Theta_{k-1}^{(i)}) }{q(\Theta_{k}^{(i)}|\Theta_{k-1}^{(i)},\mathbf{z}_{1:k})}$
		\ENDFOR
		\STATE normalise weight $w_{k}^{(i)} = \bar{w}_{k}^{(i)}/(\Sigma_{i=1}^{N}\bar{w}_{k}^{(i)})$
		\STATE calculate effective sample size $N_{eff}=1/\Sigma_{i=1}^{N}(w_{k}^{(i)})^2$
		\IF{$N_{eff}<N_{T}$} 
		\STATE resample $\{\Theta_{k}^{(i)}, \omega_{k}^{(i)}\}_{i=1}^{N}$ 
		\STATE apply a MCMC (Monte Carlo Markov Chain) move
		\ENDIF 
		\ENSURE posterior samples: $\{\Theta_{k}^{(i)}, \omega_{k}^{(i)}\}_{i=1}^{N}$
	\end{algorithmic}
	\caption{Particle filter for source parameter estimation }
	\label{alg:PF_STE}
\end{algorithm}

A similar process is employed in calculating predicted posteriors with hypothesised measurements under a candidate control action but with some simplifications to reduce online computational load.

\subsection{Implementation of optimisation}

To reduce the computational load, in this paper we choose the optimal control input from a finite set of one-step action, i.e. $\mathbf{u}_{k} \in \mathcal{U} := \{ \uparrow, \, \downarrow, \, \leftarrow, \, \rightarrow, \, \nwarrow, \nearrow, \, \swarrow, \, \searrow \}$. The step size can be determined based on the operation environment. 

The possible future measurements for a given control input $\mathbf{u}_k$ can be generated using the agent model (\ref{eq:agent}), the dispersion model (\ref{eq:advdif}) and the sensor model (\ref{eq:sensor}) with $\Theta_{k} \sim \rho_{k|k}$. Based on (\ref{eq:pf}), a set of samples to represent the distribution of $\rho(k|k)$ are generated. For each sample $\Theta^{i}$, the dispersion model (\ref{eq:advdif}) is run to generate the chemical concentration at the new agent location $\mathbf{p}_{k+1|k}$ due to the move $\mathbf{u}_k$. A set of measurements $\hat{z}_{k+1}$ are obtained by randomising the measurements using the sensor model (\ref{eq:sensor}) with the described noise characters $v_k$ and $\bar{v}_k$. The optimal control action is selected as the one minimising the cost (\ref{eq:DCEE}) or (\ref{eq:theorem1}).


\section{Extension and relationships with other methods} \label{sec:extension}

This section first extends the basic DCEE framework presented in Section~\ref{sec:CAD} in a number of directions (including the incorporation of an unknown environment). The relationships with several other methods will then be discussed. It is shown that DCEE covers the cost/reward functions in both classic MPC and IPP, and is better fitting with autonomous search and control system design for autonomous systems in general.     

\subsection{Extension}

\subsubsection{Unknown environment}
We have only considered unknown parameters related to the target such as its location in Section~\ref{sec:CAD}. In real operation, the operation environment is also more likely to be partially known. For example, we may only know wind direction and speed within a certain range from weather forecast or meteorologic data. We now extend our DCEE algorithm to cope with uncertainty in both the source and environment. Physical parameters in the dispersion model depend on the type of chemical release and the environment. We now consider that all the related parameters in the Isotropic plumemodel (\ref{eq:advdif}) are unknown but with certain prior information. More specifically, in addition to the position and release rate of the source, the parameters to be estimated during the search process also include wind direction $\phi_s$, wind speed $u_s$, the diffusivity $\zeta_{s1}$, and the average particle lifetime $\zeta_{s2}$. That is, $\Theta =[\mathbf{s},q_s, \phi_s, u_s, \zeta_{s1}, \zeta_{s2}]\mathsf{T}$. These parameters are related to the environment (e.g. wind field, temperature) and specific chemicals. It is assumed that all the parameters are unknown but constant during the search process. 

By the virtue of the Bayesian estimate framework, we are able to make use of any prior information of these unknown parameters in our estimation. This will improve the accuracy of the parameter estimation of the target. There are no major changes to the structure of Algorithm~\ref{alg:PF_STE} except the computational burden. This extended version of DCEE will be implemented and tested  in real-time search operation on a robot in Section~\ref{sec:exp}.

\subsubsection{Multistage DCEE}
For the convenience of illustrating the basic concept of DCEE, only a one-step-ahead cost function is considered in Section~\ref{sec:CAD}. It is straightforward to extend it to DCEE with a multistage cost function as
\begin{equation}
    \label{eq:DCEEM}
J(\mathbf{U}_{k}) =\mathbb{E}_{\Theta} \left[ \mathbb{E}_{\hat{z}_{k+1}, \ldots, \hat{z}_{k+N}} \left[ \| \mathbf{p}_{k+N|k} - \mathbf{s} \|^2 | \mathcal{Z}_{k+N|k} \right] \right]
 \end{equation}
where $\mathbf{U}_k=\left[\mathbf{u}_k, \ldots,\mathbf{u}_{k+N-1} \right]$, $\mathcal{Z}_{k+N|k}=\left[\mathcal{Z}_{k},  \hat{z}_{k+1}(\mathbf{p}_{k+1|k}), \ldots, \hat{z}_{k+N}(\mathbf{p}_{k+N|k}) \right]$ where $\mathbf{P}_{k+i|k}, i=1, \ldots, N$, are the predicted locations of the agent under the control sequence $\mathbf{U}_k$ with the agent model (\ref{eq:agent}). This implies that the conditional distance between the robot position and the location of the source in the next $N$th step is of interest. In other words, the predicted posterior of $N$th step with the hypothesised future $N$ step measurements under possible control sequence $\mathbf{U}_k$ is used to improve our autonomous search strategy. Then this cost function is minimised over the set of admissible control $\mathbf{U}_k$  to give the optimal control sequence $\mathbf{U}_k^*$ but only the first move $\mathbf{u}^*_k$ is implemented and the process is repeated in a receding horizon fashion.   Although the multistage strategy may improve the performance of the autonomous search further, it brings a significant increase of computation burden with it. 


\subsection{Relationship with other methods}
\subsubsection{Information theoretic approaches} \label{sec:IPP}
As briefly discussed in Section~\ref{sec:CAD}, information theoretic approaches have been used to develop autonomous search methods where the search process is treated as an information gathering problem. More specifically, the aim of motion planning is to reduce the uncertainty of the estimation of the target location and other unknown environmental factors. Therefore, the next move of the robot is determined to maximise a reward function related to information. Mutual information on entropy, Kullback–Leibler divergence and other metrics are used to measure the uncertainty of the information and the corresponding search schemes have been presented. More specifically, Entrotaxis enforces the maximum entropy sampling principle which dictates the agent moves to the position that is least certain \cite{hutchinson2018entrotaxis}.  Instead of reducing predicted entropy, infotaxis aims to minimise the predicted posterior variance of the source location \cite{ristic2016study} \cite{ristic2017autonomous}. This directly links to exploration effect of DCEE, i.e. the second term of (\ref{eq:theorem1}).

Since Entrotaxis exhibits a slightly better performance than Infotaxis \cite{hutchinson2018entrotaxis}, we implement Entrotaxis as a benchmark of IPP for comparison with DCEE in our simulation and experimental study. 

\subsubsection{Model Predictive Control (MPC)} \label{sec:MPC}

As discussed in Introduction, currently there is little work looking into autonomous search from a control engineering perspective. If following our idea of considering the search as a process to drive the agent towards the source, it could be formulated in a \emph{classic} stochastic MPC framework as
\begin{equation} \label{eq:MPC}
    J(\mathbf{u}_{k}) =\mathbb{E}_{\Theta} \left[ \| \mathbf{p}_{k+1|k} - \mathbf{s} \|^2 | \mathcal{Z}_{k} \right] 
\end{equation}
where only the measurements up to the current time are used in estimating the location of the source. A simple interpretation of this cost function is that a right strategy is to drive the robot towards the best estimate of the source location with all the information we have so far. 

In a similar fashion, this cost function could be rewritten as
\begin{equation} \label{eq:theorem1MPC}
   J(\mathbf{u}_{k})= \| \mathbf{p}_{k+1|k} - \bar{\mathbf{s}}_{k|k}\|^2 + P_{k|k}
\end{equation}
where $\bar{\mathbf{s}}_{k|k}$ is the mean estimated location of the source at time $k$ conditional on $\mathcal{Z}_{k}$, and  $P_{k|k}$ is the covariance matrix of the estimation $\mathbf{s}_{k|k}$ defined as $P_{k|k}=\mathbb{E} \left[ \tilde{\mathbf{s}}_{k|k}^T\tilde{\mathbf{s}}_{k|k} \right]$. 

Since the control $\mathbf{u}_k$ does not affect the current estimation so its uncertainty, the second term in (\ref{eq:theorem1MPC}) is not relevant. The control action only affects the future position of the agent. So broadly speaking, the cost function (\ref{eq:MPC}) in classic stochastic MPC corresponds to the exploitation effect of DCEE, i.e. the first term of (\ref{eq:theorem1}), i.e., to drive the robot to the best estimate location of the source based on all the current available information. This is not surprising since control is about to make use of information to take action. During the process of driving towards the believed source location, the robot also collects new measurements so its belief about the source location is updated at each step with these new measurements. However, this is \emph{accidental} or \emph{passive} learning to reduce the uncertainty. It shall be highlighted that recently there are works in MPC with \emph{active} learning using dual control effort that will be discussed in the next section \cite{Mes2018}. 

In summary, the MPC and information theoretic approaches can be considered as special cases of the the proposed DCEE: the former is biased to exploitation while the latter exploration. The simulation and experimental comparisons between MPC, Entrotaxis and DCEE will be presented in Section~\ref{sec:simulation} and \ref{sec:exp}. 

\subsubsection{Dual control of uncertain systems}

According to Bar-Shalom and Tse \cite{BarTse1974}, a control input is said to have dual control effect if it can affect, with nonzero probability, at least one $r$th-order central moment of a state variable ($r>1$). In a series of seminal papers, Feldaum first recognised when controlling an uncertain system, control inputs have a dual effect; i.e.  not only a directing effect on system states but also a probing effect on system uncertainty \cite{Fel1960a} \cite{Fel1960b} \cite{Fel1961a} and \cite{Fel1961b}. Most of the work in dual control is devoted to a system with unknown or immeasurable states. It has been shown in the so-called \emph{separation principle} of control engineering, that a state estimator and a controller could be designed separately does not hold for most of systems, except, i.e. Linear Quadratic Gaussian (LQG). Various works have been presented to exploit the dual effect of control inputs. However, although it is very conceptually attractive, dual control is computationally intractable even for moderately sized systems \cite{Wit2008}. Recently triggered by the advances in MPC and active learning, there is a renewed interest in dual control in the context of stochastic MPC with active learning \cite{Mes2018}. The probing effect of control inputs is investigated to actively learn the system state so to reduce the level of uncertainty in state estimation in an MPC setting.         

Another area concerning the dual effect of control signals is adaptive dual control, e.g. \cite{Wit2008}, \cite{FilUnb2000}, \cite{heirung2017dual}. Most of the adaptive control methods are developed based on the Certainty Equivalence (CE) principle where a control law is developed by treating the estimated parameters as the true parameters and a parameter updating mechanism is adopted to online estimate the unknown parameters. However, except in very special cases, control signal affects not only the output but also the quality of the parameter estimation. This separation principle implied by CE does not hold for most of adaptive control. There is also an intrinsic conflict between them since one side, information about the process increases with the level of perturbation and on the other side, the system output shall vary as little as possible. Adaptive dual control was proposed to address these issues; see \cite{Wit2008}, \cite{FilUnb2000}. 

To address the computationally intractable challenge of dual control, a number of methods have been proposed to find suboptimal solutions. Broadly speaking, there are two types of MPC with dual control effect: indirect dual control and explicit dual control. The former is to approximate the dynamic programming \cite{Ber2005} while the latter adds the probing activity by deliberately modifying the cost function. A number of suboptimal approaches have been developed, but so far there are only a few reports about the applications of adaptive dual control or MPC with active learning \cite{Mes2018}. 

Although DCEE is similar to these dual control methods that explicitly exploit the dual effects of control signals, there are clear differences between them. In the MPC with active learning or other dual control methods, the dual effect concerns the probing effect on a control system itself (i.e. unknown states or parameters). 

\emph{Rather than about a control systems itself, DCEE exploits the probing effect on unknown parameters of the tasks a control system is designed to perform (e.g. the source location in this study), or an unknown environment that a control system operates in. So it goes beyond the traditional use of dual control effects, e.g. \cite{Wit2008}, \cite{FilUnb2000}, \cite{heirung2015mpc}, \cite{heirung2017dual}, \cite{Mes2018}. DCEE aims to increase the level of autonomy, or deal with control problems where only high level specifications (goals) are defined. The control system itself has to find more information about the goals and the environment in order to perform the tasks by probing the environment}.          

\section{Simulation study} \label{sec:simulation}
To test the efficacy of the proposed solution, a simulation study is performed which compares DCEE against the competitive solutions, most notably, \acs{MPC} described in Section~\ref{sec:MPC} and IPP. It shall be highlighted that there is no existing work about applying MPC in autonomous search which itself also forms a new contribution of this paper. Surprisingly, we find that sometimes a simple MPC algorithm achieves comparable results with information theoretic approaches. For IPP, we implement a typical search algorithm, Entrotaxis proposed in \cite{hutchinson2018entrotaxis}.  In all the simulation studies and the experiment studies in the next section, we keep everything the same except the cost/reward function. By testing in this manner, the joint exploitation/exploration characteristics of DCEE can be thoroughly evaluated.






\subsection{Simulation Scenario}

The simulation is performed under the setting of a single source release within an open bounded environment. The agent deployed to search the area is a UAV. The simulated source is modelled with a constant release rate using the \acs{IP} model (\ref{eq:advdif}).
The gas sensing model (\ref{eq:sensor}) outlined in Section \ref{sec:formulation} is used for taking simulated measurements from the \acs{IP} ground truth model. To make the results meaningful, in total 360 runs of simulation have been conducted. 

Model parameters and initial prior values of $\Theta$ are outlined in Table \ref{tab:simParam} where $N(m,\delta)$ denotes a normal distribution with mean of $m$ and variance of $\delta$ while $\gamma(m,\delta)$ a gamma distribution with a shape parameter $m$ and a scale parameter $\delta$. \acs{UAV} operational parameters are provided in Table \ref{tab:UAVParam}. The search area is defined by $[x_{min},x_{max}],[y_{min},y_{max}, [z_{min},z_{max}]$ in Table~\ref{tab:simParam}  with $U$ denotes a uniform distribution as a prior of the source location. In this simulation study, we follow Section~\ref{sec:CAD} that only the location $(\mathbf{s}_x,\mathbf{s}_y,\mathbf{s}_z)$ and the release rate $q_s$ of the source are considered to be unknown.

\begin{table}[ht]
    \centering
    \caption{Model parameters for the ground truth simulated plume with corresponding estimation engine prior parameters. `source' represents one of the 12 values of source configurations}
    \label{tab:simParam}
    \begin{tabular}{lll}
        Model Parameters       & Ground Truth  & Prior used in algorithms \\
        \hline
        \hline
        $x$ position $s_x$              & source    & $U(x_{min},x_{max})$ \\
        $y$ position $s_y$              & source    & $U(y_{min},y_{max})$ \\
        $z$ position $s_z$              & $1$m      & $U(z_{min},z_{max})$ \\
        Release rate $q_s$              & $5$g/s   & $\gamma(2,5)$ \\
        Wind speed $u_s$              & $4$m/s    & $N(4,2)$ \\
        Wind direction $\phi_s$           & source    & $N(source,10)$ \\
        Diffusivity coefficient $\zeta_{s1}$   & $1$       & $N(1,2)$ \\
       Average particle lifetime $\zeta_{s2}$   & $8$       & $N(8,2)$ \\
        Number of Particles in PF       & -         & $20,000$ \\
        Threshold of sensor $h_{thr}$               & -         & 0.5 \\
    \end{tabular}
\end{table}

\begin{table}[ht]  
\caption{Operational parameters for simulated UAV}
    \label{tab:UAVParam} 
    \centering
       \begin{tabular}{ll}
        Operational Parameters  & UAV   \\
        \hline
        \hline
        Motion Model            & Linear (i.e. Eq.(\ref{eq:agent})) \\
        Velocity                & 2m/s \\
        Flight budget           & 900s\\
        Search algorithms   & DCEE, MPC, Entrotaxis \\
        Step size               & 2m \\
        Step directions          & $[0^o,45^o, \dots ,315^o]$ \\
        Predictions per step    & 40 \\
        Start Position          & $[2,2,4]$ \\
    \end{tabular}
\end{table}

\begin{figure}
    \centering
    \includegraphics[width=\columnwidth]{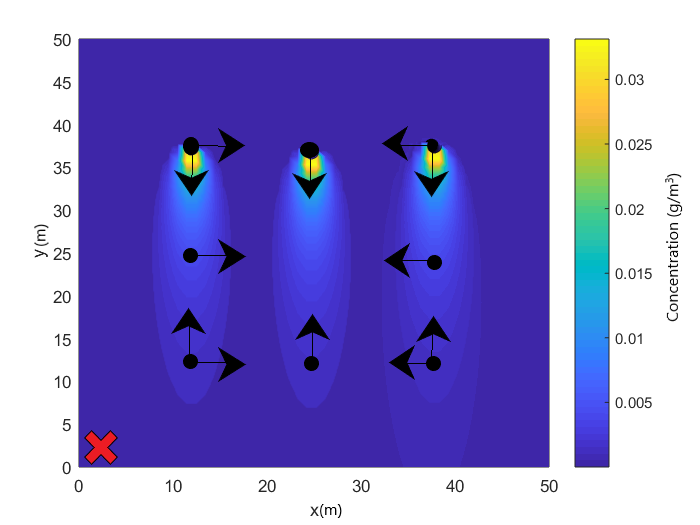}
    \caption{12 sources and their respective wind directions, as indicated by the black arrow, within the bounded search domain. Three example plumes are shown with a coloured scale. Red 'X' represents the UAV's start location.}
    \label{fig:sim12cond}
\end{figure}

To ensure the reliability of results, 12 different source configurations are tested as shown in Figure \ref{fig:sim12cond} (differing in source location $(\mathbf{s}_x,\mathbf{s}_y$) and the wind direction $\phi_s$). It is key that a selection of these source plumes do not directly overlap with a chord drawn from the \acs{UAV}s start location and $(\mathbf{s}_x,\mathbf{s}_y,\mathbf{s}_z)$. This adds the realistic element that a prior may not accurately represent the true solution and that navigating directly towards the current belief is not guaranteed to attain good data. Each source is also repeated 10 times each for a total 120 simulation runs per search method.

\subsection{Results and discussion}
\begin{figure}
    \centering
    \includegraphics[width=\columnwidth]{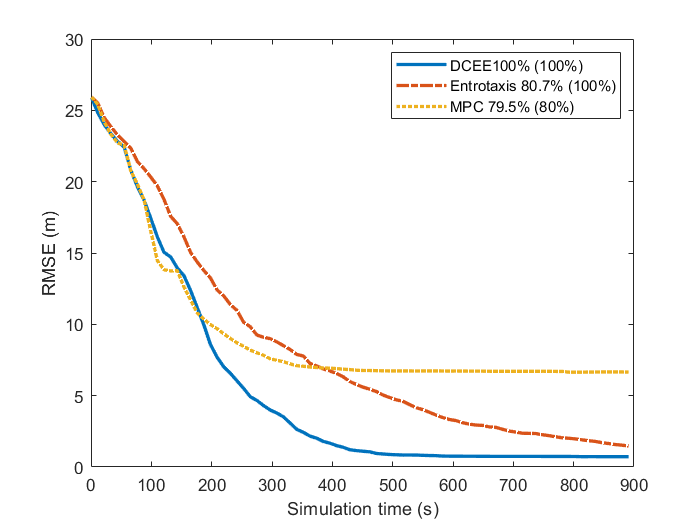}
    \caption{Average RMSE over time for all 120 simulation runs. Also pictured, each method's source acquisition rate and plume acquisition rate respectively.}
    \label{fig:simRMSE}
\end{figure}

To measure the performance of each search method over time, the \acs{RMSE} of estimated $(\mathbf{s}_x,\mathbf{s}_y,\mathbf{s}_z)$ against that of the true source is recorded at each sampling event. The average \acs{RMSE} for each method across all sources is shown in Figure \ref{fig:simRMSE}.

Figure~\ref{fig:simRMSE} shows the performance of these three strategies. At a glance, it looks that at the beginning, \acs{MPC} demonstrates comparable steepening of the \acs{RMSE} reduction gradient as DCEE. But this is partially due to the simulation setting and the location of the sources. In MPC, the location estimation is given by the mean of the probability distribution $p(\Theta_k)$, i.e. $\bar{\mathbf{s}}_{k|k}$. At the beginning, a uniform distribution of the possible location of the source is used since there is no prior information in particle filtering. Its mean is located in the middle of the search area so naturally MPC drives the UAV towards the middle of the area. When a source is placed far away from the start point of the UAV as in most of the cases (which was intended to make the search more challenging ), it gives an impression that the distance to the true location of the target reduces quickly. In other words, it accidentally occurs, not intentionally. 
Entrotaxis, however, shows a shallower gradient at the beginning due to the tendency to travel perpendicular to wind direction in order to reduce the entropy of the estimation in the particle filter. At approximately 200s, \acs{MPC} fails to continue to adequately resolve the source whereas DCEE continues gathering useful data and convergence to a \acs{RMSE} of 0.5m at approximately 500s. Entrotaxis also manages to converge to a similar accuracy but at the end of the UAV flight time at 900s.

\begin{figure*}
    \centering
    \includegraphics[width=\columnwidth]{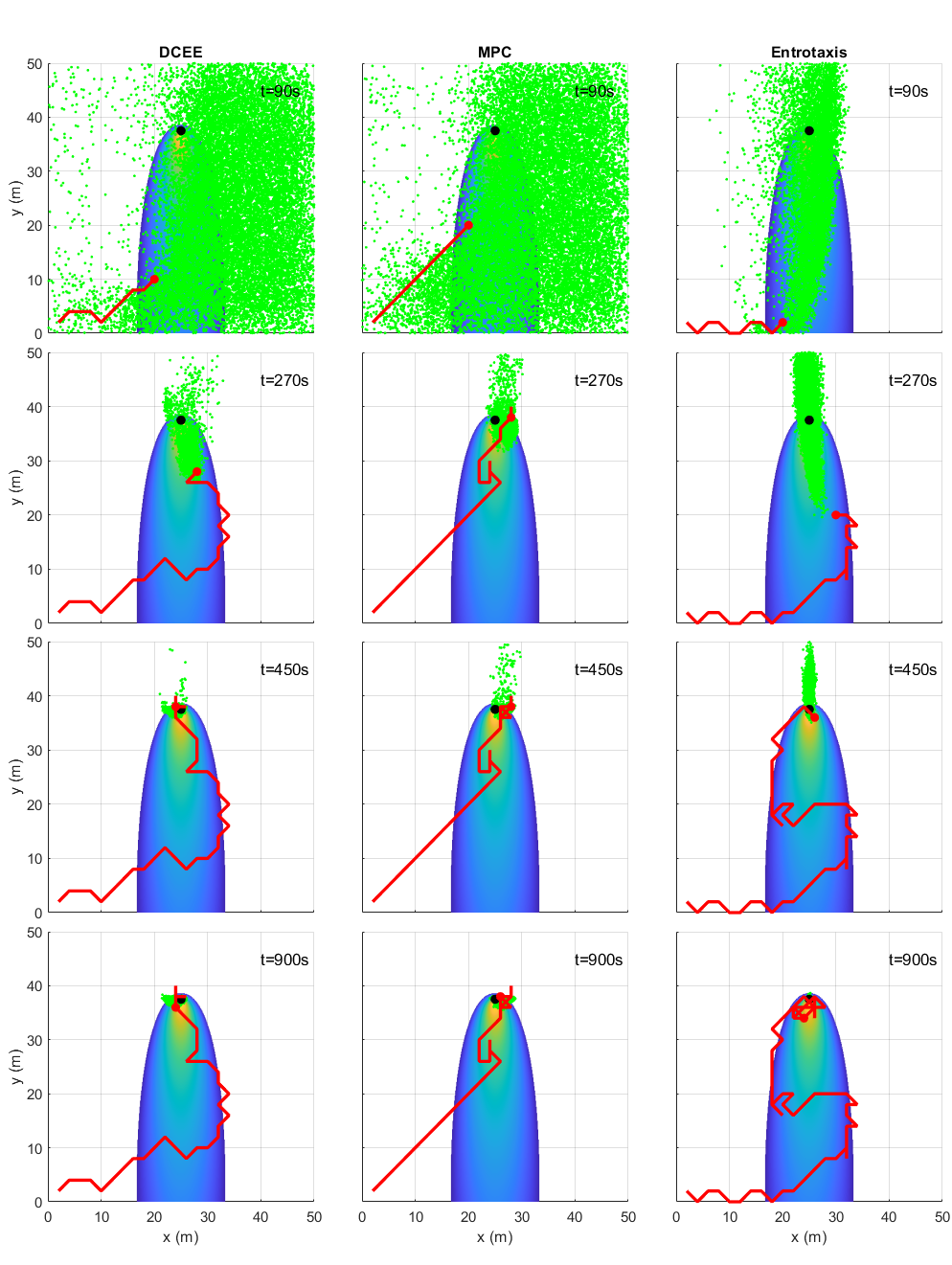}
    \caption{Comparison of source location estimation over time for each control method on a single run at 90s, 270s, 450s and 900s. Each small green dot represents a single hypothesis of the source location in the particle filter. The red dot represents the UAVs current sampling location and the red line represents the historic trajectory taken.}
    \label{fig:simComp}
\end{figure*}

Figure \ref{fig:simComp} shows a side by side comparison of each control strategy in a representative run at 90s, 270s, 450s and 900s. The behaviours described above are shown by the exemplar trajectories. But since this is a scenario where the source is in the middle of the search area, MPC shows better convergence than on average (however convergence is still noticeably worse than both DCEE and Entrotaxis).

The source acquisition rate and the plume acquisition rate are also recorded in Fig.~\ref{fig:simRMSE}. For a control method to be defined as having acquired the source, the final \acs{RMSE} of an individual run must be below $3$m. The \acs{UAV} is deemed to have acquired the plume when there is a sampling event yielding a reading greater than the minimum threshold $h_{thr}$ of the sensor model during the search process. These are both important metrics to consider during evaluation as although performance is primarily dictated by an ability to reduce the distance to the true source quickly, the success rate of finding the source and plume gives a greater insight as to operational qualities of each control method. DCEE achieves $100\%$ success in the source acquisition rate in all 120 simulations, which is much better than MPC and Entrotaxis that have similar successful rate. Both DCEE and Entrotaxis achieves $100\%$ in plume acquisition but MPC only achieves $80\%$. Entrotaxis is unable to find the source location before the flight budget depletes in $80.7\%$ of cases. \acs{MPC} is only able to find the plume in $80\%$ of cases, but is able to resolve the source in $99\%$ of those cases. This lack of plume acquisition explains why an average convergence \acs{RMSE} of 7m is seen for the \acs{MPC} strategy.

\begin{figure*}
    \centering
    \includegraphics[width=\columnwidth]{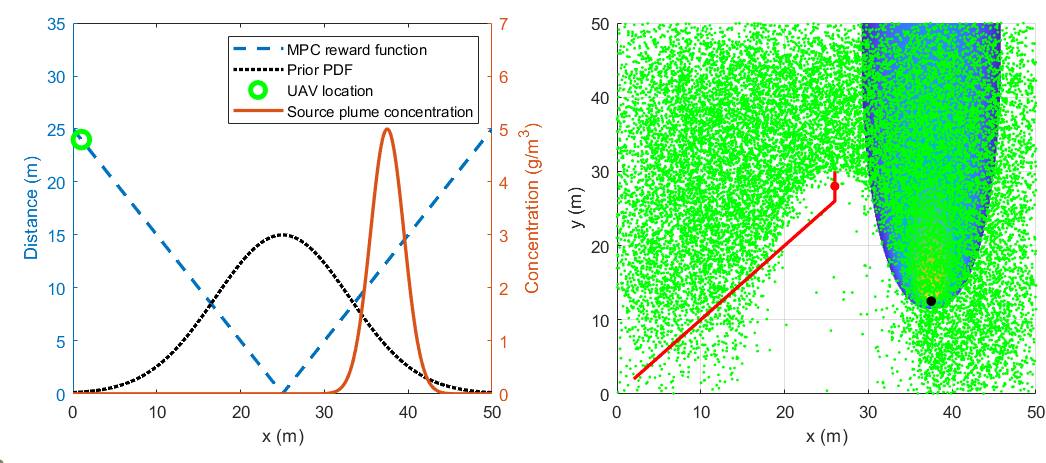}
    \caption{Exemplar behaviour of \acs{MPC} in failed search cases. 1 dimensional example (left) and simulation example (right). Simulation conditions shown are $[\mathbf{s}_x,\mathbf{s}_y,\phi_s]=[37.5m,12.5m,0^o]$ and the UAV reached the displayed control point at t=260s, and continued to sample between the same two points until the end of simulation at t=900s.}
    \label{fig:MPCminima}
\end{figure*}

The lack of plume acquisition for \acs{MPC} may be caused by local minimum during the search, or a large mismatch between the prior information of $(\mathbf{s}_x,\mathbf{s}_y,\mathbf{s}_z)$ and the ground truth. This can be shown by studying an one-dimensional case of \acs{MPC} source term estimation shown in Figure \ref{fig:MPCminima}. In this figure, the \acs{MPC} reward function (blue dashed line) dictates the control strategy given the prior PDF condition (black dotted line). Along this trajectory, no information can be gained due to the mismatch between the prior PDF and the plume, and therefore the UAV will reach a local minimum from which it cannot escape. 

It is clear that DCEE outperform these two methods significantly in all aspects. This is because DCEE is able to generate the best autonomous search strategy in trading off between exploration and exploitation in all the scenarios. It shall be mentioned that DCEE and Entrotaxis has a similar computational burden but MPC has a much less computational demand than these two other methods.

\section{Experimental Study with a ground robot} \label{sec:exp}

\begin{figure}
    \centering
    \includegraphics[width=\columnwidth]{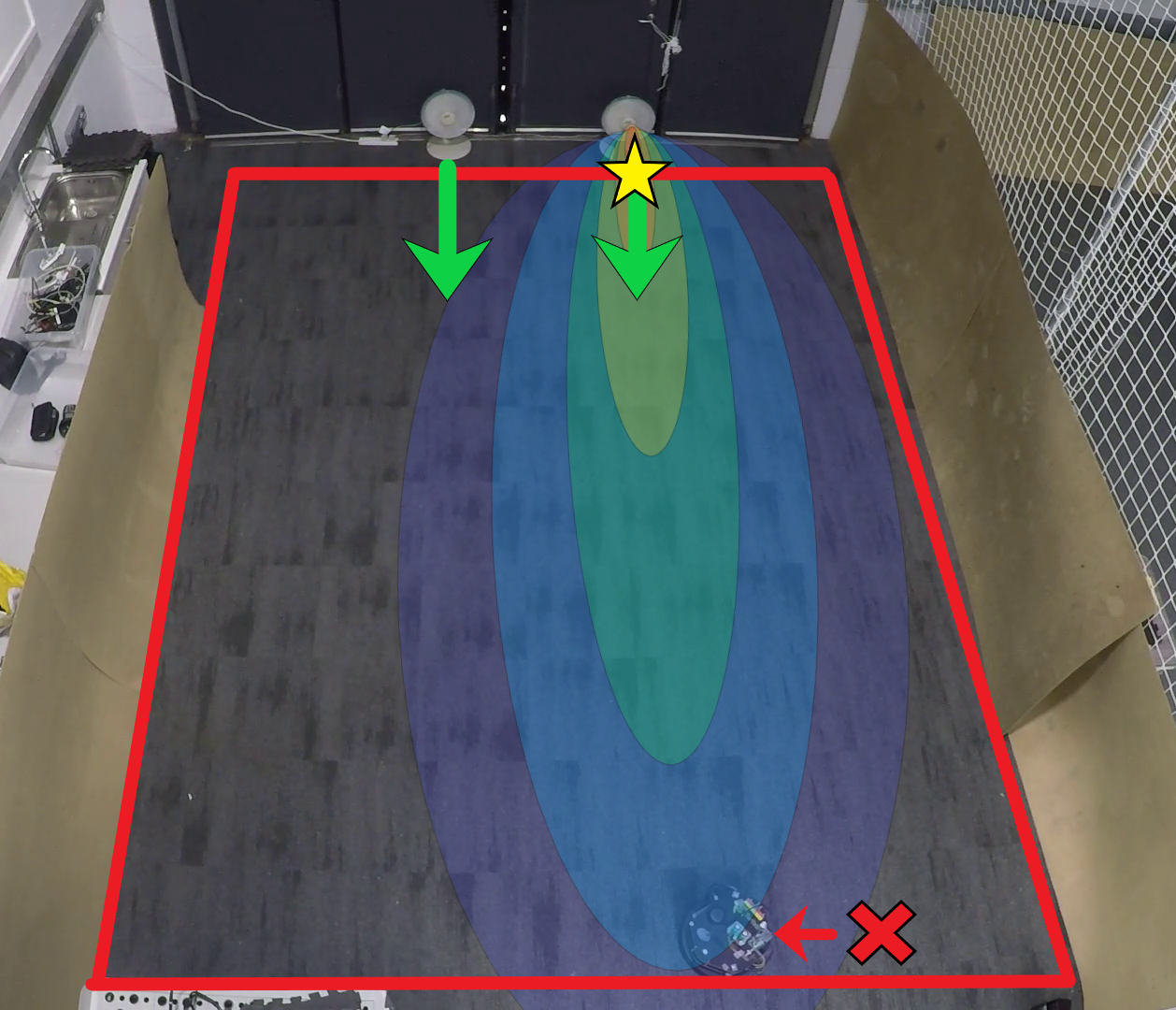}
    \caption{Experimental scenario. Red boundary shows the prior search boundary limits, green arrows show the wind field source and direction, the yellow star marks the acetone source location and the red X marks the downwind start location for the search. 
    }
    \label{fig:expScenario}
\end{figure}

\begin{figure}
    \centering
    \includegraphics[width=10cm]{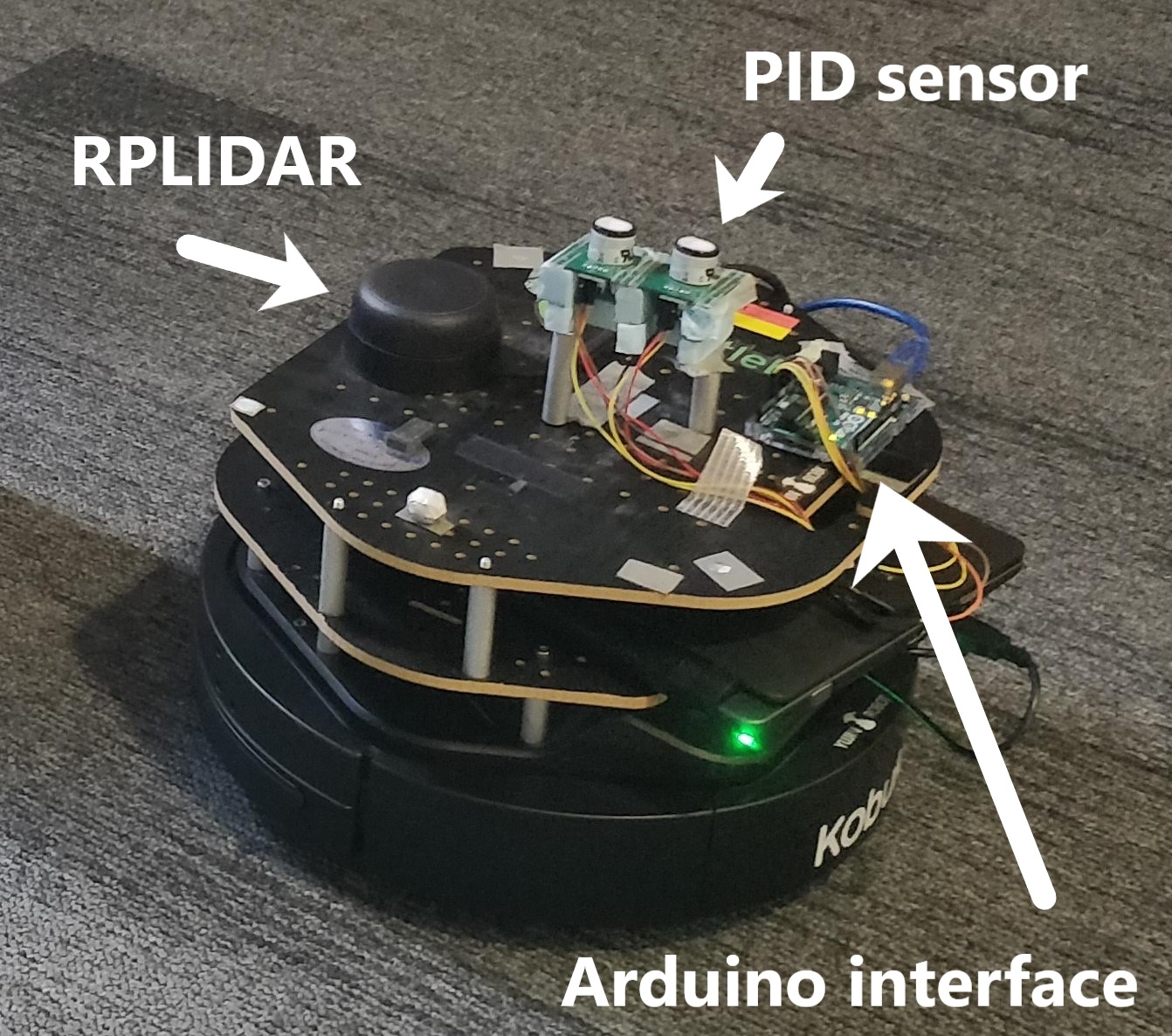}
    \caption{Turtlebot 2e with 2 Alphasense PIDs, Arduino interface and $360^o$ RPLIDAR A2.}
    \label{fig:robot}
\end{figure}

\begin{figure}
    \centering
    \includegraphics[width=\columnwidth]{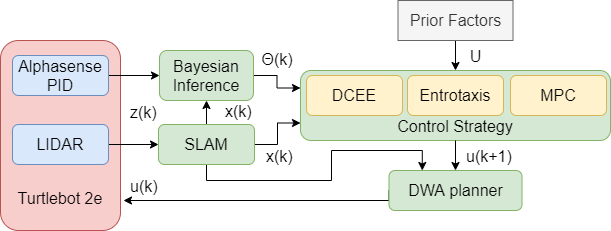}
    \caption{Data flow for the experimental testing of autonomous search algorithms using a ground robot}
    \label{fig:expFlow}
\end{figure}

To further validate the findings in the simulation, these three autonomous search algorithms are implemented in an indoor environment using a ground robot. They are tested and evaluated through a large number of experiments (in total 60 runs). The same 3 control methods from the simulation study are compared and are largely unchanged in their setup (changes are only made in appropriate operational parameters due to the size of the search area and the agent model from UAV of the third order to the ground robot of a second order). However, one major difference between the simulation and experiment tests is that the the parameters associated with both the sources and the environment are now treated as unknown and their estimation uncertainty is considered through dual control effect, i.e. the algorithm presented in Section~\ref{sec:extension} is applied. More specifically, the parameters estimated online in the search process consist of the source location ($\mathbf{s}_x$, $\mathbf{s}_y$), the release rate $q_s$, wind direction $\phi_s$, wind speed $u_s$, the diffusivity $\zeta_{s1}$, and the average particle lifetime $\zeta_{s2}$. The source of the airborne release used experiments is a small beaker of acetone, which is agitated by the air flow of two fans that create a wind field of $\sim1.5$m/s across the test area. A fresh air intake as well as exhaust fans are used to ensure that the plume is stable over the 60 runs for a fair comparison. Figure \ref{fig:expScenario} shows the 4.5x3.5m area to be searched. The same motion step constraints of 1 step ahead and 8 directions are applied, with a step size of 0.3m. 
A ground robot (Turtlebot 2e) equipped with a RPLIDAR A2 laser scanner and 2 alphasense photoionisation detectors (PID) are used to collect concentration measurements of the acetone vapour (Figure \ref{fig:robot}). Since one of the aims of the experimental study is to closely recreate real-world deployment, it is also not assumed that mapping, localisation and low-level control/planner are known or perfect. Mapping and localisation are performed using the popular \ac{SLAM} technique of Hector SLAM \cite{kohlbrecher2011flexible}, whilst low-level planner is achieved by using a \ac{DWA} algorithm \cite{fox1997dynamic}. Data flow for the experimental setup is shown in Figure \ref{fig:expFlow}. 

The main functional difference between the simulation and experiments is that there is greater model uncertainty between the dispersion model used in Bayesian estimation (and the search algorithms) and the real dispersion event. Whereas the estimation algorithm could guarantee convergence in the simulations (due to using the same plume model, certain localisation and perfect control), the source parameters in the experimental scenario are unknown and values have to be estimated during the autonomous search. 
These factors make the online estimation aspect much more challenging. Furthermore, since the DCEE strategy is deeply coupled with the Bayesian estimation to decide the best next move based on the level of estimation uncertainty, it is important to investigate the performance of DCEE under dispersion model mismatching and unmodelled dynamics in real applications. Given these factors, plus the new uncertainty in localisation, mapping and low-level planning/control actions, the experiment provides a challenging stage upon which any weaknesses of DCEE will become apparent. Successful application in the experimental trial will also show that the DCEE framework is suitable for real-world deployment on autonomous search systems.

Since the indoor experiment proposed is relatively constricted in the number of configurations that can be tested, two start locations of the robot are be explored. The first (as pictured in Figure \ref{fig:expScenario}) is downwind of the source but close to the plume. The second is situated in the top left corner of Figure \ref{fig:expScenario}, and therefore upwind but far across from the plume. Testing these two extremes will show the relative merits and demerits of each control method and also test the versatility of DCEE. Each location is tested 10 times per control method (60 total runs).

\subsection{Downwind Results}
\begin{figure}
    \centering
    \includegraphics[width=\columnwidth]{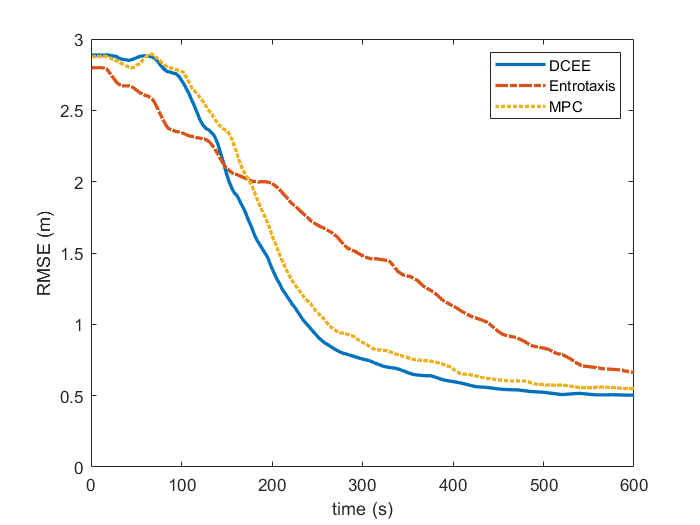}
    \caption{Average RMSE over time for each control method during the downwind start experimental testing.}
    \label{fig:expRMSE1}
\end{figure}

After performing all runs, the \acs{RMSE} over time is recorded for each algorithm and averaged across the 10 repeats. The data is shown in Figure \ref{fig:expRMSE1}. The graph shows that Entrotaxis has the quickest initial reduction of error but is overall slower than both DCEE and \acs{MPC} to converge. \acs{MPC} and DCEE show very similar error reduction behaviour with modest gains (approximately 60s quicker convergence) seen when using DCEE.

For the downwind start (as shown in Figure \ref{fig:expScenario}), the ground robot is very close to the edge of the plume and is therefore somewhat guaranteed to move into a position where it can collect information. Due to the almost immediate guarantee of data, exploitation methods are inherently favoured to reduce source term uncertainty. \acs{MPC} finds the plume and then travels directly upwind thus reducing longitudinal uncertainty quicker (which being the most uncertain direction manifests in a quicker overall error reduction). This explains why MPC exhibits a surprising performance in this setting.   

Entrotaxis travels primarily perpendicular to the wind direction and therefore establishes the lateral extent of the plume quickly but then more slowly converges to an accurate location. DCEE follows a similar trajectory to \acs{MPC} but exhibits more lateral movements across and upwind of the plume. Since these lateral movements are small relative to the length scale of the plume, this is only shown as a modest error reduction over \acs{MPC}. These tendencies also explain the superior rate of tracking error reduction of both \acs{MPC} and DCEE for this particular starting location. Each method successfully located the source location for all runs. 

    
\subsection{Upwind Results}
\begin{figure}
    \centering
    \includegraphics[width=\columnwidth]{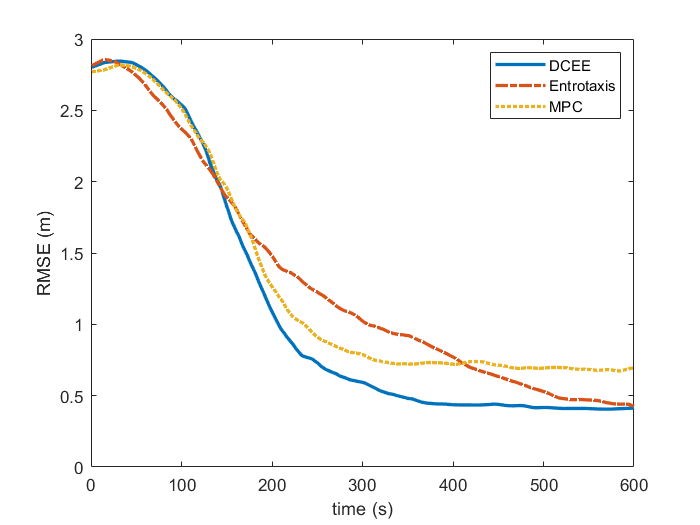}
    \caption{Average RMSE over time for each control method during the upwind start experimental testing.}
    \label{fig:expRMSE2}
\end{figure}

The data from the upwind tests are shown in Figure \ref{fig:expRMSE2}. Similar trends to the downwind test are seen but Entrotaxis shows much more competitive performance and both Entrotaxis and DCEE have a significantly better convergent \acs{RMSE} than \acs{MPC}.

For the second starting scenario, the starting location is far outside the plume boundary and the robot has to explore to gather concentration readings. Therefore unlike the first starting location, there is non-detection at the first several steps. So this is a more challenging situation and more close to real life. 

For Entrotaxis, slow perpendicular travel to the wind field is exhibited which is favourable for the second starting location due to its longitudinal proximity to the source location. This results in a much fast error reduction compared to the downwind scenario.

Due to non-detection at the beginning, \acs{MPC} navigates towards the centre of the search space and catches the edge of the plume. The robot is then navigated towards the source but since the robot is longitudinally close to the source location, only a few data points are collected before the robot reaches the source. At this stage, \acs{MPC} struggles to converge the model to a more accurate location as it is stuck at a minimum near to the source and has not collected enough data along its trajectory.

Similarly to the downwind test, DCEE takes a more diagonal trajectory and is closer longitudinally to the source when the robot records a concentration reading. Unlike \acs{MPC}, when DCEE navigates the robot to the source location at the boundary, further lateral operations are performed and the robot is therefore capable of converging to an accurate estimate similarly to Entrotaxis.

This second test has therefore shown that in a scenario where exploration has superior terminal accuracy, DCEE is capable of showing both the convergent accuracy as well as the error reduction rates of the exploitation based strategies.

Overall, the experimental tests have further validated the results seen in the simulation study. They have also shown that the proposed DCEE technique has credible application in a real-world scenario, as well as outperforming techniques previously deployed for autonomous search. Different from simulation, MPC shows a competitive performance in experiments but this is mainly due to the small scale and restriction of the lab environment.    

\section{Conclusion}  \label{sec:conclusion}
For an autonomous search problem, a control theoretic approach is proposed in this paper.  Inspired by the dual control concept, DCEE is able to take into account both exploitation and exploration effect of control/decision/planning actions. It is suitable for designing high level control systems for a system operating in an unknown environment and (or) with unknown parameters associated with a task (i.e. unknown location of a source in this paper). Simulation and experiment results show this new framework outperforms classic MPC, a popular method in control engineering, and IPP, an information theory approach widely used in robotics and autonomous systems. For the specific autonomous search of hazardous sources, it is shown that DCEE maintains an optimal balance (in the sense of the principle of optimality (Bellman, \cite{Bellman1957Dynamic})) between the probing activity and control activity of control inputs, which are naturally in conflict. In the DCEE framework, stochastic MPC, active learning, IPP, exploitation and exploration, and autonomous search are closely related and integrated. Much more work is required in further exploring and exploiting this new framework.

The level of autonomy of a system can be measured in terms of the level of complexity of tasks it is able to perform and the level of uncertainty it is able to cope with \cite{antsaklis2020autonomy}. Hence it could be argued that any autonomous system shall be able to perform certain tasks and achieve defined goals based on its belief, but also update its belief and reduce the level of uncertainty by actively exploring the environment. Consequently any decision making/planning/control action shall have (dual) effect in these two aspects. The DCEE framework proposed in this paper provides a promising vehicle in realising these desirable properties for autonomous systems. It will inspire more research into developing similar goal-oriented control systems.   

\begin{ack}
This work was supported by the UK Engineering and Physical Sciences Research Council (EPSRC) Established Career Fellowship ``Goal-Oriented Control Systems: Disturbance, Uncertainty and Constarints'' under the grant number EP/T005734/1.
\end{ack}

\bibliography{TCSTbibfile}
\end{document}